%% file: acl_latex.tex
\pdfoutput=1

\documentclass[11pt]{article}

\usepackage[final]{acl}
\usepackage{CJKutf8}

\usepackage{times}
\usepackage{latexsym}
\usepackage{float}
\usepackage[T1]{fontenc}

\usepackage[utf8]{inputenc}

\usepackage{microtype}

\usepackage{inconsolata}
\usepackage{enumitem}
\usepackage{graphicx}
\usepackage{booktabs} 
\usepackage{multirow}
\usepackage{color}
\usepackage{amsmath}
\usepackage{xcolor}
\usepackage{colortbl}
\usepackage{xspace}

\definecolor{deepgreen}{RGB}{0,128,128}

\newcommand{\red}[1]{\textcolor{red}{#1}}
\newcommand{\green}[1]{\textcolor{green}{#1}}
\newcommand{\blue}[1]{\textcolor{blue}{#1}}

\title{Mitigating Hallucinations of Large Language Models in Medical Information Extraction via Contrastive Decoding}

\author{Derong Xu\textsuperscript{1,2}, Ziheng Zhang\textsuperscript{3}, Zhihong Zhu\textsuperscript{4}, Zhenxi Lin\textsuperscript{3}, Qidong Liu\textsuperscript{2},\\ {\bf Xian Wu\textsuperscript{3}\footnotemark[1], Tong Xu\textsuperscript{1}\thanks{Corresponding authors.}, Xiangyu Zhao\textsuperscript{2}\footnotemark[1], Yefeng Zheng\textsuperscript{3,5}, Enhong Chen\textsuperscript{1}\footnotemark[1]}\\
        \textsuperscript{1}University of Science and Technology of China  \& State Key Laboratory of\\ Cognitive Intelligence, \textsuperscript{2}City University of Hong Kong, \\ \textsuperscript{3}Jarvis Research Center, Tencent YouTu Lab,
        \textsuperscript{4}Peking University, \textsuperscript{5}Westlake University \\
derongxu@mail.ustc.edu.cn, \{tongxu, cheneh\}@ustc.edu.cn, \\\{zihengzhang, chalerislin, kevinxwu, yefengzheng\}@tencent.com, \\\{qidongliu2-c, xianzhao\}@cityu.edu.hk, zhihongzhu@stu.pku.edu.cn 
}

\begin{document}
\maketitle
\begin{abstract}
The impressive capabilities of large language models (LLMs) have attracted extensive interests of applying LLMs to medical field. However, the complex nature of clinical environments presents significant hallucination challenges for LLMs, hindering their widespread adoption. In this paper, we address these hallucination issues in the context of Medical Information Extraction (MIE) tasks by introducing ALternate Contrastive Decoding (ALCD).
We begin by redefining MIE tasks as an \textit{identify-and-classify} process. We then separate the identification and classification functions of LLMs by selectively masking the optimization of tokens during fine-tuning. During the inference stage, we alternately contrast output distributions derived from sub-task models. This approach aims to selectively enhance the identification and classification capabilities while minimizing the influence of other inherent abilities in LLMs. 
Additionally, we propose an alternate adaptive constraint strategy to more effectively adjust the scale and scope of contrastive tokens. Through comprehensive experiments on two different backbones and six diverse medical information extraction tasks, ALCD demonstrates significant improvements in resolving hallucination issues compared to conventional decoding methods.

\end{abstract}
\input{section/intro}
\input{section/relatedwork}
\input{section/method}

\input{section/experiment}
\input{section/conclusion}

\bibliography{acl_latex}

\appendix
\input{section/appendix}

\end{document}

%% file: section/intro.tex
\section{Introduction} \label{sec:intro}

Medical Information Extraction (MIE), including tasks such as medical entity recognition and relation extraction, is a fundamental component of medical NLP~\cite{hahn2020medical,xu2024editing}. It enables the derivation of structured knowledge from plain text, benefiting a wide array of applications \cite{wang2024llm4msr,liuyue_KGE_SymCL,qi2024unimel}, like medical knowledge graph construction~\cite{wu2023medical,xu2024multi,xu2023multimodal}, medical dialogue~\cite{gao-etal-2023-dialogue-medical,wu2024medkp}, and medical report generation~\cite{liu2021exploring}. Previous MIE tasks~\cite{biobert,guan2020cmeie} have been supervised, and their performance heavily depends on the quality and quantity of available training data. However, labeling medical documents requires specific knowledge which is both costly and time-consuming.

\begin{figure}[t]
\centering
\includegraphics[width=1\linewidth]{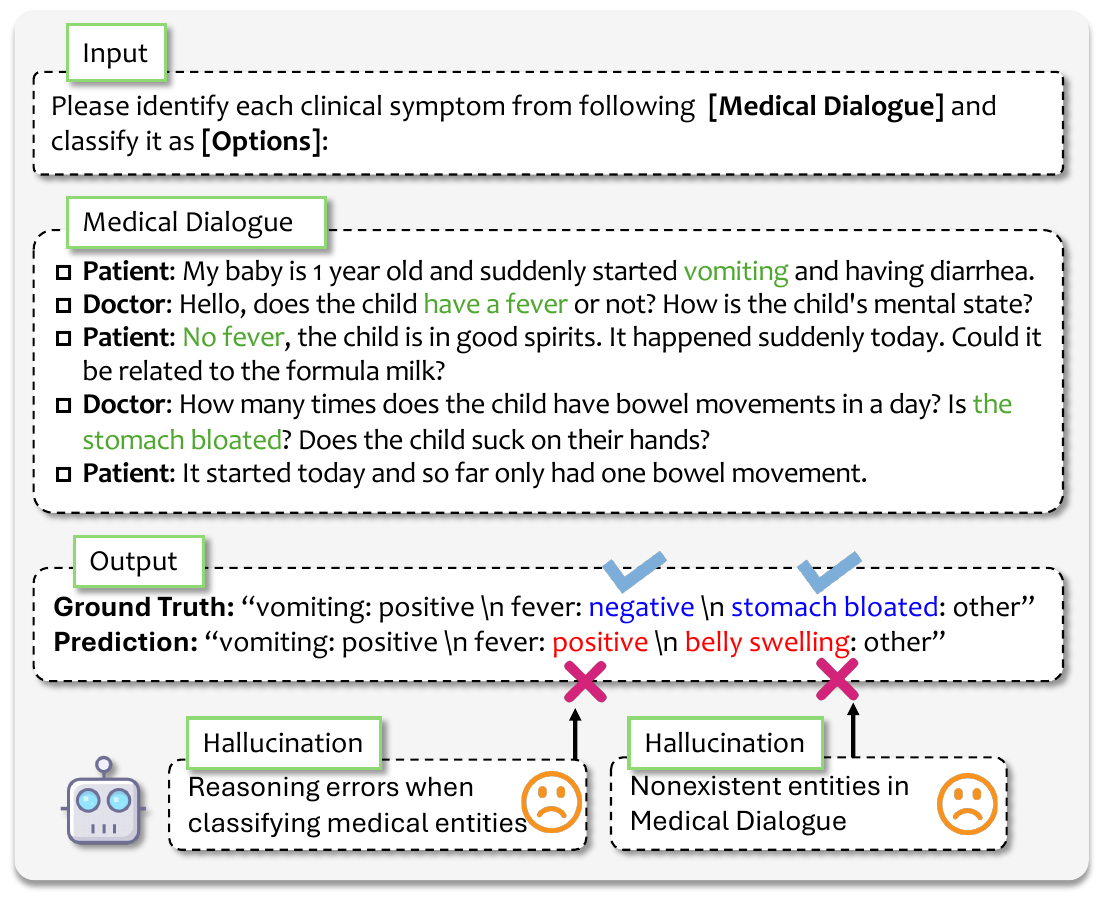}
\caption{An example demonstrating the hallucination generated by LLMs in MIE tasks. The \green{green} font in medical dialogue indicates a high correlation with ground truth. The \blue{blue} font in the output represents correct token, while the \red{red} font represents tokens with hallucination problems. These problems mainly include the presence of nonexistent entities and reasoning errors.
}
\label{fig:intro}
\end{figure}

Recently, the remarkable zero-shot capabilities of large language models (LLMs) such as ChatGPT and GPT-4 \cite{openai2023gpt4} have inspired researchers to transform MIE tasks into a generation paradigm~\cite{zhu2023promptcblue}. However, the medical domain is less tolerant of errors compared to other domains. While there have been attempts to apply LLMs to the medical field~\cite{medpalm1,usmle,liu2024moe,liu2024largedistilling}, there is a growing concern about the issue of hallucination~\cite{huang2023survey}. In the context of MIE, two types of hallucinations exist: (1) LLMs may identify medical entities that are not present in original texts, thereby fabricating facts and deviating from the original information. (2) LLMs may face reasoning errors when classifying medical entities, due to statistic biases in the pre-trained corpus. We show such a hallucination problem in Figure \ref{fig:intro}.

In this paper, we address the challenges of hallucination when applying LLMs to MIE tasks. We observe that LLMs for MIE can be conceptualized as an \textit{identify-and-classify} process: initially identifying potential medical concept spans from the plain text, and then classifying these text spans into predefined categories (e.g., start token of a specific entity, subject of a specific relation), as shown in the `Output' of Figure \ref{fig:intro}.
The natural approach to applying LLMs is to prompt them to simultaneously complete both \textit{identify} and \textit{classify} steps in a unified decoding process \cite{lu2022unified,wang2023instructuie}.
 We speculate that the hallucination problem may be linked to the joint next-word generation abilities of identification and classification, which could have inadvertently compromised each other's performance.
Therefore, we believe that decoupling abilities of identification and classification, allowing LLMs to concentrate on specific sub-tasks, could simplify the complexity of the MIE task and potentially reduce hallucination issues~\cite{khot2022decomposed,bian2023inspire}.


Motivated by the aforementioned observation, we introduce ALternate Contrastive Decoding (ALCD), a straightforward decoding strategy designed to enhance the performance of LLMs on MIE tasks. In the training stage, we mask the optimization of tokens separately to decouple the identification and classification models. For instance, when fine-tuning the parameters of the identification model, classification tokens are masked to focus the model's attention solely on identification tokens, thereby ignoring its classification capability. During the inference stage, ALCD bolsters its classification/identification ability and contrasts logit predictions with another model. This contrastive decoding process alternates between classification and identification, depending on the type of the next token, which is determined by a simple rule-based judgment. Furthermore, we propose an adaptive constraint strategy to dynamically adjust the scale and scope of contrastive tokens. This allows individual samples to adapt to their unique characteristics by measuring the consistency among the three models and the level of confidence.
Overall, this work makes three key contributions:

\begin{itemize}[topsep=0pt, partopsep=0pt, leftmargin=13pt, parsep=0pt, itemsep=3pt]
\item To our knowledge, we are the first to employ contrastive decoding as a strategy to reduce hallucinations in LLMs for MIE tasks.
\item We validate the broad applicability of our ALCD approach through experiments using two LLM backbones across six diverse medical tasks, such as determining causal relationships in medical concepts~\cite{zhu2023promptcblue}. \footnote{https://github.com/quqxui/quqxui-AlternateCD}
\item Our experimental results underscore the superiority of ALCD over eight established decoding methods. 
\end{itemize}


%% file: section/relatedwork.tex
\section{Related Work} \label{sec:realtedwork}

\subsection{LLMs for Medical Domain}


Rapid development has been seen in directly employing general LLMs (e.g., ChatGPT~\cite{openai2023gpt4}, ChatGLM~\cite{chatglm}, and Qwen~\cite{qwen}) to the medical domain and training medical LLMs using medical data, such as Med-PaLM~\cite{medpalm1}, clinicalGPT~\cite{clinicalgpt}, and MedAlpaca~\cite{medalpaca}. Both general LLMs and medical LLMs may suffer from hallucinations, the undesired phenomenon of LLMs generating contents not based on training data or facts when applying them to complex medical tasks.
Hallucinations could be caused by multiple factors, such as imperfect representation learning or erroneous decoding~\cite{hall_survey}. Due to the high demand for reliability in the medical domain, the hallucinations are thus less tolerated. Although previous works have explored the problem of hallucination in the medical domain \cite{umapathi2023med,ji-etal-2023-towards}, there is a lack of exploration in MIE task, particularly regarding the efficiency of different decoding methods for mitigating hallucination.

\subsection{Contrastive Decoding}
The idea of contrastive decoding for LLM has been explored in various previous works, and different decoding strategies focus on different aspects of LLM improvements.
Contrastive Decoding (CD)~\cite{CD} is proposed to contrast output probability of large-scale expert LLMs with small-scale amateur LLMs to diminish undesired amateur behavior and improve fluency and coherence in the generated contents.
Context-aware Decoding (CAD)~\cite{CAD} focuses on the issue of LLMs' insufficient attention to context. CAD downweights output probability associated with LLMs' prior knowledge to promote LLMs' attention to context, thus improving the faithfulness of the generated contents.
\citet{DOLA} introduced DoLa, where the output next-word probability is obtained from the difference in logits between a higher layer versus a lower layer, to reduce hallucinations and enhance truthfulness in the knowledge-based question-answering tasks.
Visual Contrastive Decoding (VCD) is another decoding method to mitigate object hallucinations for large vision-language models by contrasting output distributions from original and distorted visual inputs~\cite{VCD}.
\citet{stayCFG} adapted Classifier-Free Guidance (CFG)~\cite{CFG} from text-to-image generation to text-to-text generation and they showed CFG can increase the LLMs' performance and adherence to various prompts, including basic prompting, chain-of-thought prompting, and chatbot prompting.

Although previous contrastive decoding strategies have been shown effective in addressing specific hallucinations in LLMs, their performance is inadequate for MIE tasks. In contrast, our ALCD effectively decouples the abilities to contrast and decode outputs, leading to notable enhancements.

%% file: section/method.tex
\section{Methodology} \label{sec:method}
In this section, we introduce ALternate Contrastive Decoding (ALCD), a method specifically designed for medical information extraction tasks. Section \ref{sec:preliminary} provides the foundational knowledge of Contrastive Decoding, while Section \ref{sec:alcd} delves into the details of our proposed ALCD method.
\subsection{Preliminary} \label{sec:preliminary}
For generative LLMs, the common method for text generation is to predict next token in an auto-regressive manner. Specifically, we denote the parameters of an LLM as $\theta$. The model utilizes input text $\boldsymbol{x}$ and system instructions (prompts) $\boldsymbol{i}$ to generate a response $\boldsymbol{y}$. For each time step $t$, we have:
\begin{equation}
\setlength{\abovedisplayskip}{5pt}
\setlength{\belowdisplayskip}{5pt}
\begin{aligned}
    y_t \sim & \mathcal{P}_\theta (y_t | \boldsymbol{i}, \boldsymbol{x}, \boldsymbol{y}_{<t}), \\
       \sim & softmax(logit_\theta (y_t | \boldsymbol{i}, \boldsymbol{x}, \boldsymbol{y}_{<t}) ),
\end{aligned}
\end{equation}
where $y_t$ represents the output token at a specific time step $t$, and $\boldsymbol{y}_{<t}$ denotes the sequence of generated token sequence until the time step $t-1$. The common ways of the next token selection include selecting the highest probability token (greedy search), exploring multiple high-probability paths simultaneously (beam search), or sampling according to the probability distribution (e.g., nucleus sampling \cite{nucleussampling}).

While, in contrastive decoding, there are typically two logits, which may be obtained from different LLMs using the same input source~\cite{CD} or the same LLM using different input sources~\cite{CAD}. 
It should be noted that they need to share the same tokenizer to keep consistency between different logits. The probability for the next token is adjusted through subtraction:
\begin{equation}
\setlength{\abovedisplayskip}{5pt}
\setlength{\belowdisplayskip}{5pt}
logit_{\theta}(y_t | \boldsymbol{i}, \boldsymbol{x}, \boldsymbol{y}_{<t}) - logit_{\theta{'}} (y_t | \boldsymbol{i}, \boldsymbol{x}, \boldsymbol{y}_{<t}).
\end{equation}
The $logit_{\theta{}}$ and $logit_{\theta{'}}$ are usually generated from an LLM with high capabilities and low capabilities, respectively. For example, in CD~\cite{CD}, $logit_{\theta{}}$ comes from a large expert LLM and $logit_{\theta{'}}$ comes from a small amateur LLM. Subtracting these two logits helps amplify the ground-truth tokens in $logit_{\theta{}}$ and downplay hallucinated tokens in $logit_{\theta{'}}$. Inspired by CD, we propose to alternately amplify or downplay the classification and identification capabilities of LLMs during the decoding process, to improve final generation results.

\begin{figure*}[t]
\centering
\includegraphics[width=1\linewidth]{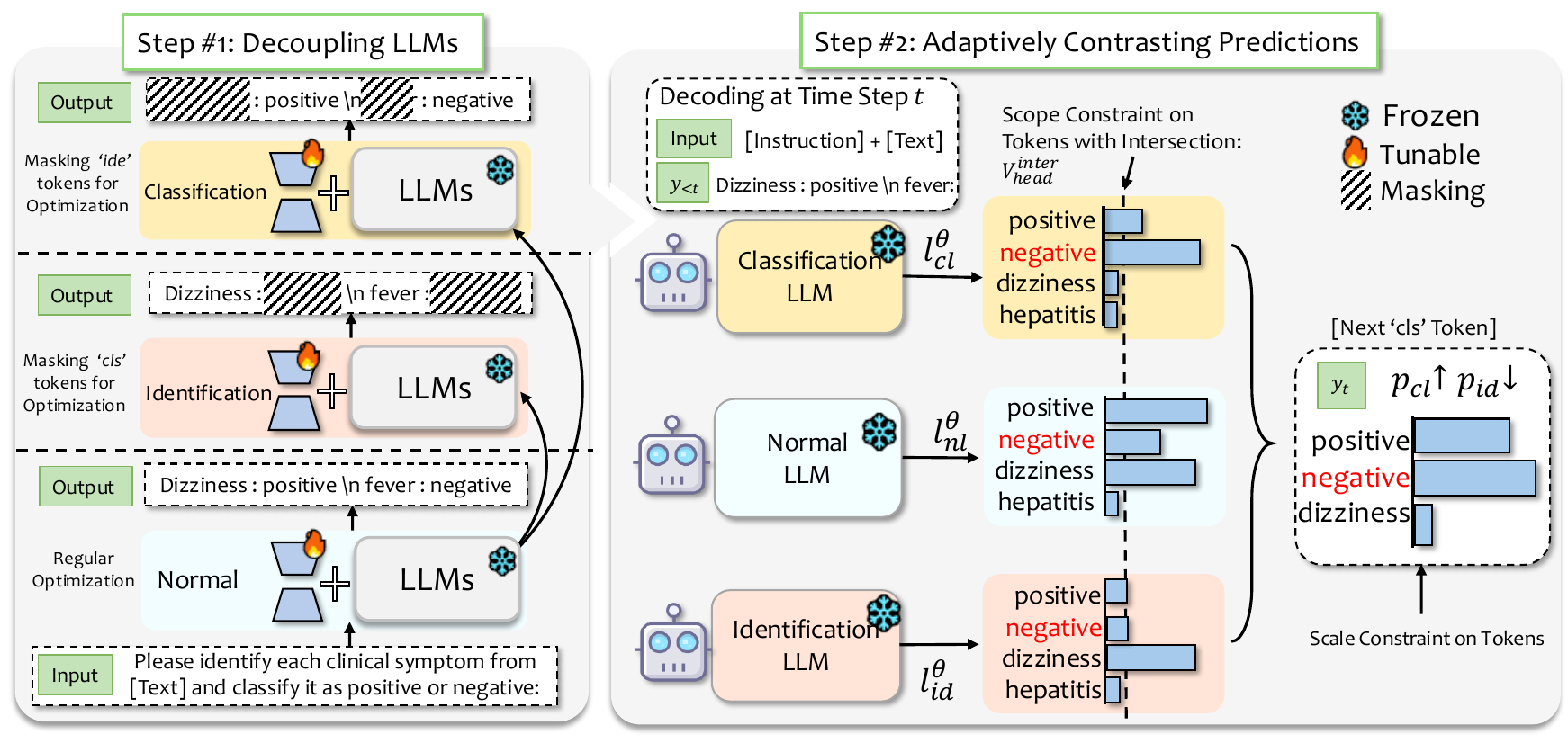}
\caption{The overall pipeline of our proposed ALCD consists of two main steps.  In Step \#1, we aim to fine-tune sub-models individually to decouple the abilities of identification and classification. In Step \#2, we adaptively contrast the predictions at each time step by applying scale and scope constraints on tokens. The figure shows how LLMs generate token $y_t$ at time step $t$ based on previous tokens $\boldsymbol{y}_{<t}$. The terms $cls, ide, other$ represent classification, identification, and other tokens, respectively. The output logits of normal, classification and identification models are represented as $l^{\theta}_{nl}$, $l^{\theta}_{cl}$, and $l^{\theta}_{id}$.}
\label{fig:method}
\end{figure*}

\subsection{Alternate Contrastive Decoding}
\label{sec:alcd}
The process of our proposed ALCD is illustrated in Figure \ref{fig:method}. We break down medical information extraction into two stages: identification and classification. In Section \ref{sec:decouple}, we fine-tune LLMs separately for identification and classification. In Section \ref{decode}, we utilize the decoders of three LLMs (identification, classification, and normal) together to perform MIE. As the two new LLMs are trained with Lora~\cite{hu2021lora}, they do not cause an excessive increase in training.

\subsubsection{Decoupling with optimization masking}
\label{sec:decouple}
To effectively harness identification and classification capabilities of LLMs while minimizing interference from one another, we propose to decompose their respective abilities. Typically, it is natural to fine-tune two subtasks independently, resulting in a identification model $\mathcal{M}_{id}$ and a classification model $\mathcal{M}_{cl}$. But this method has distinct instructions and input-output formats compared to normal model $\mathcal{M}_{nl}$. It poses an issue when these models are combined during the inference step, which can lead to inconsistent input with fine-tuning step, ultimately reducing the accuracy.

In this work, we propose to optimize two capabilities separately using optimization masking during the fine-tuning process, as shown in Figure \ref{fig:method}(Step \#1).
We employ the same inputs as original task for fine-tuning both $\mathcal{M}_{id}$ and $\mathcal{M}_{cl}$ models.
During fine-tuning, we selectively optimize tokens, and for instance, when optimizing parameter $\theta_{id}$ of identification model $\mathcal{M}_{id}$, we mask the tokens for classification task:
\begin{equation}
\setlength{\abovedisplayskip}{5pt}
\setlength{\belowdisplayskip}{5pt}
\scalebox{0.95}{$
\mathop{\max}\limits_{\theta_{id}} \sum\limits_{(\boldsymbol{x},\boldsymbol{y})\in \mathcal{D}} \sum\limits^{|y|}_{t=1, t \notin \mathcal{T}_{cl}} log(\mathcal{P}_{\theta_{id}} (y_t | \boldsymbol{i}, \boldsymbol{x}, \boldsymbol{y}_{<t})), 
$}\label{lora_id}
\end{equation}
where $\mathcal{T}_{cl}$ represents the time step of classification tokens, which do not require optimization, and $\mathcal{D}$ denotes training dataset. On the other hand, when optimizing parameter $\theta_{cl}$ of classification model $\mathcal{M}_{cl}$, we mask the tokens for identification task:
\begin{equation}
\setlength{\abovedisplayskip}{5pt}
\setlength{\belowdisplayskip}{5pt}
\scalebox{0.95}{$
\mathop{\max}\limits_{\theta_{cl}} \sum\limits_{(\boldsymbol{x},\boldsymbol{y})\in \mathcal{D}} \sum\limits^{|y|}_{t=1, t \notin \mathcal{T}_{id}} log(\mathcal{P}_{\theta_{cl}} (y_t | \boldsymbol{i}, \boldsymbol{x}, \boldsymbol{y}_{<t})), 
$}\label{lora_cl}
\end{equation}
where $\mathcal{T}_{id}$ represents time step of identification tokens. By employing masking optimization, we expect to develop LLMs that possess diverse capabilities. For fine-tuning normal model $\mathcal{M}_{nl}$, we also employ formulas similar to \ref{lora_id} and \ref{lora_cl}, but without any masking operations. Given the constraints of computational resources, we implemented parameter-efficient fine-tuning techniques (e.g., LoRA~\cite{hu2021lora} ) to train these models.

\subsubsection{Adaptively Contrasting the Predictions}
\label{decode}
After decoupling the capabilities, a significant challenge arises: how can we effectively harness the individual abilities of sub-models? To address this, ALCD is designed to alternate the enhancement of the classification ability of $\mathcal{M}_{cl}$ and the identification ability of $\mathcal{M}_{id}$ during LLM's inference stage, while excluding the influence of other capabilities originally present in normal model $\mathcal{M}_{nl}$. An illustration is shown in Figure \ref{fig:method}(Step \#2).

We denote $n_t \in \{cls, ide, other\}$ as the type of next token prediction, where $cls, ide, other$ indicate classification, identification, and other tokens, respectively. Generally, in order to facilitate the evaluation of text generated from LLMs, it is typically to present the output of MIE in a structured format~\cite{lu2022unified}. Therefore, when LLMs generate token $y_t$ at time $t$, we can determine the next token based on previous tokens $\boldsymbol{y}_{<t}$ using a simple rule-based judgment: In our case, we require LLMs to utilize colon `:' to split $ide$ and $cls$ tokens, and each $ide$-$cls$ pair is separated by a newline character `$\backslash$n'. For instance, in this text: \textit{``Dizziness: positive$\backslash$n fever: negative''}, the $ide$ tokens (\textit{Dizziness} or \textit{fever}) are expected to be followed by a colon and then a $cls$ token (\textit{positive} or \textit{positive}). 
 We abbreviate the representation $logit_{\theta}( \cdot | \boldsymbol{i}, \boldsymbol{x}, \boldsymbol{y}_{<t})$ generated by  $\mathcal{M}_{nl}$,  $\mathcal{M}_{cl}$, and  $\mathcal{M}_{id}$ as $l^{\theta}_{nl}$, $l^{\theta}_{cl}$, and $l^{\theta}_{id}$, respectively. The overall formula is as follows:
\begin{equation}
\setlength{\abovedisplayskip}{5pt}
\setlength{\belowdisplayskip}{5pt}
\begin{aligned}
&l^{\theta}_{nl} + \alpha (d_{cl}*l^{\theta}_{cl} - d_{id} * l^{\theta}_{id}), {\rm if\ } n_t = cls \\
&l^{\theta}_{nl} + \alpha ( d_{id} * l^{\theta}_{id} - d_{cl} * l^{\theta}_{cl}), {\rm if\ } n_t = ide \\
&l^{\theta}_{nl}, \qquad  \qquad  \qquad \qquad \quad {\rm if\ } n_t = other
\label{equation:Contrasting the Predictions}
\end{aligned}
\end{equation}


where $\alpha$ is a hyper-parameter and analyzed in Section \ref{sec:Hyper-parameters Scale}. 
$d_{id}$ and $d_{cl}$ are adaptive scales proposed to measure the distance between two logit distributions: one between $\mathcal{M}_{nl}$ and $\mathcal{M}_{id}$, and the other between $\mathcal{M}_{nl}$ and $\mathcal{M}_{cl}$. We leverages Jensen-Shannon Divergence (JSD) to calculate them:
\begin{equation}
\setlength{\abovedisplayskip}{5pt}
\setlength{\belowdisplayskip}{5pt}
\begin{aligned}
d_{id} &= JSD(logit_{\theta_{nl}}|| logit_{\theta_{id}}), \\
d_{cl} &= JSD(logit_{\theta_{nl}}|| logit_{\theta_{cl}}). \label{equation:jsd}
\end{aligned} 
\end{equation}
Specifically, when predicting the next token in Formula (\ref{equation:Contrasting the Predictions}), ALCD includes two extra components in addition to the logit $l^{\theta}_{nl}$ of the normal model. For example, if $n_t$ is a $cls$ token, The first component is enhancing $l^{\theta}_{cl}$, with the motivation to utilize the classification ability of sub-model $\mathcal{M}_{cl}$. If the outputs of $\mathcal{M}_{cl}$ is more different from $\mathcal{M}_{nl}$ (e.g., larger $d_{cl}$), we will be more inclined towards the classification model.
The second component involves contrasting the influence of sub-models  $\mathcal{M}_{id}$, by decreasing logit values $l^{\theta}_{id}$ through adaptive scales ($d_{id}$). The motivation behind this is that if the outputs of $\mathcal{M}_{id}$ is more different from $\mathcal{M}_{nl}$ (e.g., larger $d_{id}$), indicating a stronger contrast (denoted as $- d_{id} * l^{\theta}_{id}$), which makes sure that ALCD has the potential to mitigate the hallucinations arising from identification ability.


Conversely, when the next token is an $ide$ token, the same rule is applicable.
For the next token that do not belong to either $ide$ or $cls$, we solely utilize logit output $l^{\theta}_{nl}$ of normal model. 
By employing this alternating contrast prediction, ALCD has the capability to modify the overall probability of tokens and then harness the abilities of sub-models.

\subsubsection{Scope Constraints on Tokens} \label{sec:Constraints}
In addition, it is worth noting that certain tokens may exhibit a significant discrepancy when subjected to contrastive decoding, which makes the implausible tokens receive a high score after contrast, leading to what is referred to as the false positives~\cite{CD, DOLA}. In light of this, we implement a constraint that is contingent upon the confidence level:
\begin{equation}
\setlength{\abovedisplayskip}{5pt}
\setlength{\belowdisplayskip}{5pt}
\begin{aligned}
&\mathcal{V}_{head}(\boldsymbol{y}_{<t})= \{v \in \mathcal{V}: \\
&\mathcal{P}_\theta (v | \boldsymbol{i}, \boldsymbol{x}, \boldsymbol{y}_{<t}) \geq \beta \mathop{\max}_{v} \mathcal{P}_\theta (v | \boldsymbol{i}, \boldsymbol{x}, \boldsymbol{y}_{<t})\},
\end{aligned}
\label{equation:constraints}
\end{equation}
where $\mathcal{V}$ represents the output vocabulary of LLMs, $v$ is the token of output vocabulary, and $\beta$ is a hyper-parameter used to determine the max truncation rate of low-probability tokens. Instead of employing constraints with a single model in \citet{CD}, our approach involves combining the intersection of confidence values $\mathcal{V}_{head}^{inter}$ obtained from three models (outputs of $\mathcal{M}_{nl}$, $\mathcal{M}_{id}$, and $\mathcal{M}_{cl}$). Tokens with confidence levels below a specific threshold are assigned a negative infinity value:
\begin{equation}
\setlength{\abovedisplayskip}{5pt}
\setlength{\belowdisplayskip}{5pt}
\scalebox{0.9}{$
\begin{aligned}
&\mathcal{V}_{head}^{inter} = \mathcal{V}_{head}^{nl} \cap \mathcal{V}_{head}^{cl} \cap \mathcal{V}_{head}^{id},\\
& logit_\theta (v | \boldsymbol{i}, \boldsymbol{x}, \boldsymbol{y}_{<t})=-\infty , \  {\rm if\ } v \notin \mathcal{V}_{head}^{inter}(\boldsymbol{y}_{<t}).
\end{aligned}
$}
\end{equation}
By combining token constraints to enhance and contrast predictions, our proposed ALCD is able to effectively leverage capabilities of $\mathcal{M}_{id}$ or $\mathcal{M}_{cl}$, while addressing the issue of hallucinations in $\mathcal{M}_{nl}$ that arise from other capabilities in $\mathcal{M}_{cl}$ or $\mathcal{M}_{id}$.

%% file: section/experiment.tex
\section{Experiments} \label{sec:experiment}

\subsection{Experimental Setup} \label{sec:experiment_setup}

\begin{table}[t]
\small
\centering
            \fontsize{9}{11}\selectfont
\setlength{\tabcolsep}{3mm}{
\begin{tabular}{lrrr}
\toprule
Dataset & \#Train &\#Valid &\#Test   \\
\midrule 
CMeEE-V2     & 4,600  & 400 & 400\\
CMeIE-V2     & 4,600  & 400 & 400\\
IMCS-V2-NER  & 4,600  & 400 & 400\\
CMedCausal   & 2,600  & 400 & 400\\
IMCS-V2-SR 	 & 4,600  & 400 & 400\\
CHIP-MDCFNPC & 4,600  & 400 & 400\\
\bottomrule 
\end{tabular}}
\caption{Dataset partitioning statistics.}
\label{tab:Statistics}
\end{table}

\begin{table*}[h]
\centering
            \fontsize{9}{12}\selectfont
\setlength{\tabcolsep}{1mm}{
\begin{tabular}{l   cccccc }
\toprule
Decoding Method & { CMeEE-V2} & {CMeIE-V2} & {IMCS-V2-NER} & { CMedCausal} & { IMCS-V2-SR} & { CHIP-MDCFNPC} \\
\midrule
\multicolumn{7}{c}{\textit{ChatGLM-6B}} \\
\midrule
Greedy Search &   66.48 & 45.60 & 88.37 & 41.01 & 71.55 & 42.58 \\
Beam Search &     66.77 & 45.80 & 88.60 & 41.41 & 71.84 & 42.77 \\
Top K Sample &    63.38 & 39.02 & 88.19 & 39.41 & 69.40 & 38.87 \\
Nucleus Sample &  64.93 & 41.13 & 88.26 & 40.58 & 69.88 & 41.92 \\
CFG{~\cite{stayCFG}} & 66.95 & 43.84 & 88.76 & 40.61 & 72.06 & 42.49 \\
CAD{~\cite{CAD}} &     66.88 & 44.04 & 88.77 & 40.57 & 72.06 & 42.49 \\
CD{~\cite{CD}} &       66.34 & 46.03 & 88.54 & 40.72 & 72.40 & 42.33 \\ 
DoLa{~\cite{DOLA}} &   66.46 & 43.78 & 88.96 & 40.47 & 38.68 & 42.92 \\
\midrule
\rowcolor{gray!10} ALCD (Ours)& \textbf{67.45}{\Large *} & \textbf{46.83}{\Large *} & \textbf{89.49}&  \textbf{42.28}{\Large *} &\textbf{73.01 }{\Large *} &\textbf{43.71}{\Large *}\\
\midrule
\multicolumn{7}{c}{\textit{Qwen-7B-Chat}} \\
\midrule
Greedy Search &  65.49 & 42.87 & 88.65 & 30.10 & 71.28 & 40.61 \\
Beam Search &    66.61 & 43.40 & 89.46 & 30.21 & 71.35 & 40.94 \\
Top K Sample &   65.71 & 36.34 & 88.83 & 19.55 & 71.04 & 40.19 \\
Nucleus Sample & 66.04 & 33.87 & 89.08 & 25.81 & 70.09 & 39.40 \\
CFG{~\cite{stayCFG}} & 65.18 & 39.07 & 88.64 & 12.96 & 71.15 & 40.18 \\
CAD{~\cite{CAD}} &     66.09 & 36.67 & 88.00 & 14.40 & 71.72 & 39.49 \\
CD{~\cite{CD}} &       65.19 & 35.86 & 88.98 & 14.69 & 70.27 & 39.35 \\
DoLa{~\cite{DOLA}} &   65.16 & 35.51 & 88.49 & 16.52 & 71.29 & 39.37 \\
\midrule
\rowcolor{gray!10} ALCD (Ours) & \textbf{67.91}{\Large *} &\textbf{44.19}{\Large *} &\textbf{90.88}{\Large *} & \textbf{31.57}{\Large *} &\textbf{72.14}{\Large *} &\textbf{ 41.88} \\
\bottomrule
\end{tabular}}
\caption{Experiment results (micro F1 score$\uparrow$: higher is better) on six medical datasets with the best scores highlighted \textbf{in bold}. All baselines are based on the same fine-tuned normal model, and the model-agnostic parameters for fine-tuning and inference are kept consistent, with only the specific decoding method being changed. ``\textbf{{\Large *}}'' indicates the statistically significant improvements (i.e., two-sided t-test with $p<0.05$) over the best baseline.}
\label{tab:mainresults}
\end{table*}

\noindent
\textbf{Tasks and Datasets.}
We apply six MIE tasks from a Chinese medical dataset named PromptCBLUE~\cite{zhu2023promptcblue} for evaluation.  \textbf{CMeEE-V2} is a task of Chinese medical entity recognition. \textbf{IMCS-V2-SR} aims to normalize the patient-doctor dialogue by medical concepts. \textbf{IMCS-V2-NER} targets extracting medical concepts from dialogues. \textbf{CMedCausal} is a task of causal relation extraction for medical texts. \textbf{CHIP-MDCFNPC} refers to clinical concept finding and discrimination. \textbf{CMeIE-V2} aims to recognize and categorize the entity relation contained in medical texts. The output forms of all tasks are built with the \textit{identify-and-classify} pattern, as mentioned in Section \ref{sec:intro}. Due to space limitations, we leave more details about the tasks to \textbf{Appendix}~\ref{appendix:task}.
Since the open-source test set was not available, we used the validation set as our test set. Subsequently, we partition the training set into a new training set and validation set and ensure the validation set contains the same number of samples as the test set.
 Table~\ref{tab:Statistics} presents the dataset partitioning statistics. 

\noindent
\textbf{Models and Baselines.}
To improve the learning of data, we experimented with two widely-used multilingual LLMs, ChatGLM-6B v1~\cite{chatglm} and Qwen-7B-Chat v1~\cite{qwen}. We compared our method for mitigating hallucinations with eight decoding baselines, which can be categorized as follows: \textbf{Deterministic decoding}: 1) greedy search decoding; 2) beam search decoding; \textbf{Stochastic decoding}: 3) Top K sample decoding; 4) nucleus sample decoding; \textbf{Contrastive decoding}: 5) CFG~\cite{CFG}; 6) CAD~\cite{CAD}; 7) CD~\cite{CD}; 8) DoLa~\cite{DOLA}. For the validation of Deterministic and Stochastic methods, we utilized the implementation provided by the Huggingface toolkit \cite{wolf-etal-2020-transformers}. However, for the contrastive decoding methods, adjustments were required when applying them to MIE tasks as they were not specifically designed to tackle the hallucination problem in MIE. For CFG, we simply use logits with normal input text and logits with the last token of input text as a comparison.
For CAD, we employ both normal input text and input text without classification labels to contrast the output in different contexts. For CD, we employ the normal model as the expert model and proceed with a model using only half the number of fine-tuning steps for the amateur model. DoLa is implemented following their published paper.

\noindent
\textbf{Implementation Details.}
We conducted all experiments using four NVIDIA V100 GPUs. As we fine-tuned LLMs using LoRA, the decoding process was performed using a single GPU. All experimental results were evaluated using the Micro-F1 score following~\citet{zhu2023promptcblue}. For ALCD, we conducted a search in the validation set to determine the appropriate values for the scale of contrasting prediction $\alpha$, the maximum rate of constraint $\beta$, and the step of fine-tuning. For $\alpha$, we limit the search scope to the values of [0.01, 0.1, 0.2, 0.3, 0.4, 0.5]. For $\beta$,  we limit the search scope to the values of [0.4, 0.45, 0.5, 0.55, 0.6, 0.65]. The fine-tuning step of the normal model remains consistent across all baselines. We employ a batch size of 8 and perform 1,000 steps to fine-tune all datasets and LLMs, except for Qwen-7B-Chat where we use 3,000 steps in CMeIE-V2, CMedCausal, and CHIP-MDCFNPC, due to that extra steps are required for convergence.

\begin{figure}[t]
\centering
\includegraphics[width=1\linewidth]{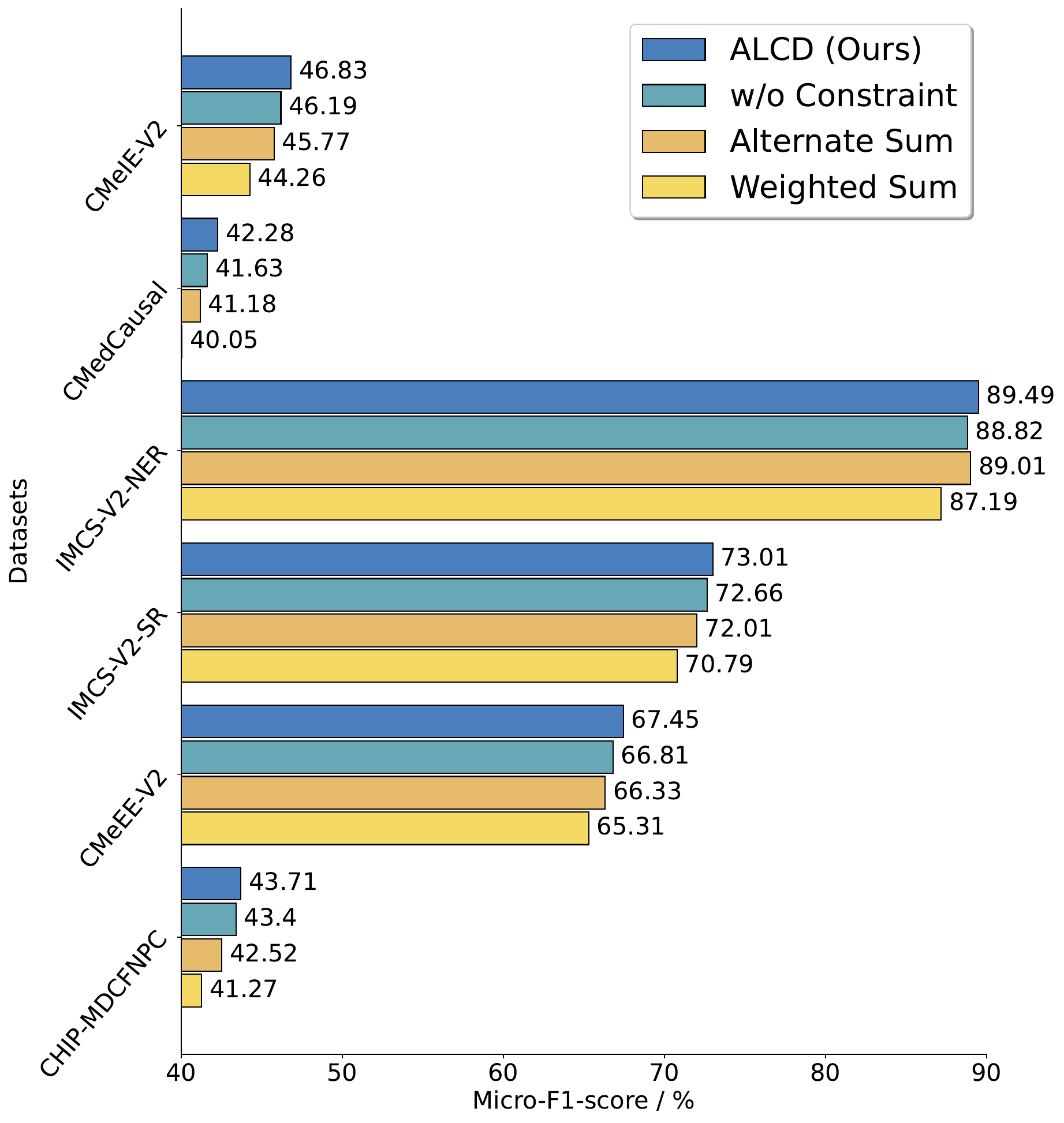}
\caption{Ablation study on six medical datasets using ChatGLM-6B.
}
\label{fig:ablationstudy}
\end{figure}

\subsection{Main Results} \label{sec:Main Results}

In this section, we provide a comprehensive performance comparison of ALCD against other baselines on six medical datasets and two different backbone LLMs.
As shown in Table~\ref{tab:mainresults}, our proposed ALCD outperforms both contrastive decoding and non-contrastive decoding methods and the performance gap reaches the largest of 4.87\% in Qwen-7B-chat on the CMedCausal dataset.
Our proposed ALCD has been shown to improve performance on both ChatGLM-6B and Qwen-7B-Chat, which confirms its universality.
Besides, ALCD particularly performs well on CMeEE-V2, IMCS-V2-NER, and CHIP-MDCFNPC datasets, and outperforms other baselines by a large margin.
This finding aligns with our motivation as these datasets include more entity candidates, more classification labels, and thus higher difficulties for LLMs.
Some contrastive decoding methods, such as DoLa, achieve much lower results on IMCS-V2-SR in the ChatGLM-6B, indicating the coupled difficulties for the medical \textit{identify-and-classify} tasks. We find that the proposed adaptive method of DoLa predominantly selects the 2nd or 8th layer as the optimal premature layer, which suggests that DoLa's intended ability to amplify factual knowledge across different layers may not be fully aligned with the MIE tasks.
We observed that the poor performance of sampling methods (Top K and Nucleus Sample) indicates that high diversity generation may not be essential for the MIE task.

\subsection{Ablation Study} \label{sec:Ablation Study}

\begin{figure}[t]
    \centering
    \includegraphics[width=0.98\columnwidth]{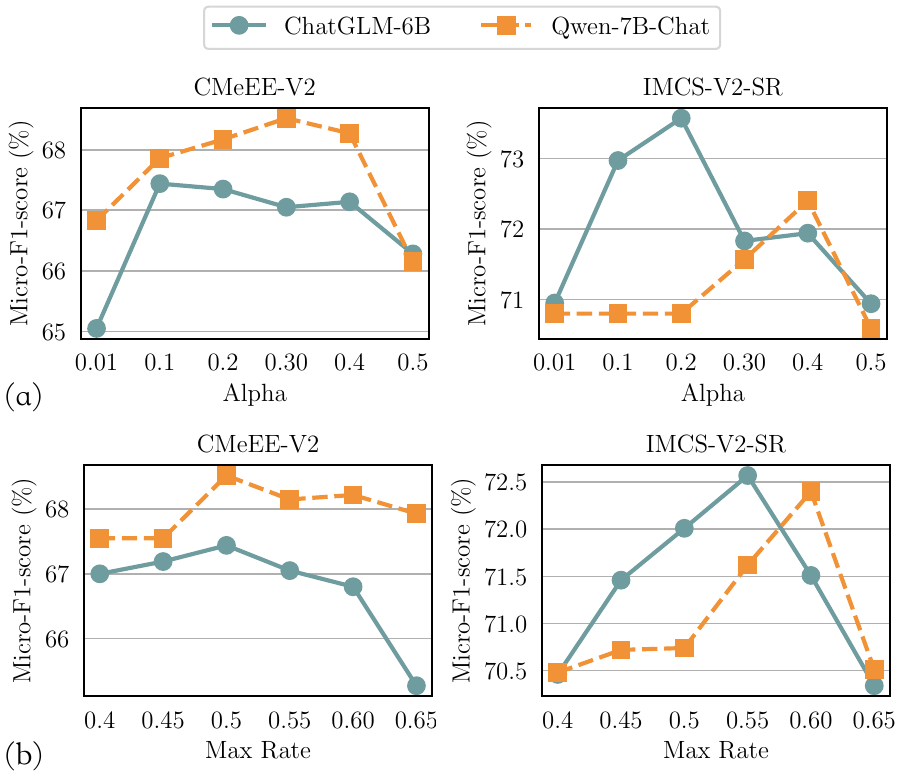}
    \caption{(a) Analysis of the scale of contrasting prediction $\alpha$ (in Formula~\ref{equation:Contrasting the Predictions}); (b) Analysis of max rate of constraint $\beta$ (in Formula~\ref{equation:constraints}).}
    \label{fig:zhexian}
\end{figure}

In this section, we analyze the effects of different components on ALCD. Specifically, we experiment with ALCD against three variants: 1) ALCD without Constraint: removing the dynamic constraints on tokens, 2) Alternate Sum: alternately summing the logits from three models instead of utilizing contrastive decoding, 3) Weighted Sum: directly summing the logits from three models with the same weight of ALCD.
As depicted in Figure~\ref{fig:ablationstudy}, the results confirm that incorporating token constraints enhances the performance of the normal model. Specifically, on the CMeIE-V2 dataset, the micro F1 score decreased from 47.02\% to 46.19\% when no constraints were utilized. Moreover, removing the alternate contrasting with either Alternate Sum or Weighted Sum resulted in performance declines, with Weighted Sum yielding the poorest overall performance.
This finding highlights the effectiveness of applying alternate contrastive decoding and indicates that solely ensembling multiple LLMs for these tasks does not lead to performance improvement.

\subsection{Scale of Contrasting Prediction} \label{sec:Hyper-parameters Scale}

To investigate the effect of hyper-parameter $\alpha$ in Formula~\ref{equation:Contrasting the Predictions}, we set different values from 0.01 to 0.5 and conduct experiments on CMeEE-V2 and IMCS-V2-SR datasets. A larger $\alpha$ means a larger scale of contrastive decoding.
As shown in Figure~\ref{fig:zhexian}(a), it can be observed that increasing the scale of contrastive decoding appropriately enhances the micro F1 score of both backbone LLMs, indicating the efficiency of our contrastive decoding method.  While, excessively large values of $\alpha$ (e.g., exceeding 0.4), can lead to a decline in performance, which demonstrates that excessive utilization or weakening of the sub-models' ability may result in a decrease in the final effect.

\subsection{Max Rate of Constraint} \label{sec:Hyper-parameters Constraint}
In this section, we examine the effect of $\beta$ in Formula~\ref{equation:constraints}, which controls the max truncation rate of low-probability tokens for contrastive decoding.
The results are shown in Figure~\ref{fig:zhexian}(b).
We observed that small $\beta$ values (e.g., smaller than 0.45) have a minimal impact on the low-probability tokens, suggesting that these tokens are unlikely to significantly influence the model. 
We also found that the performance reaches its peak at around 0.5 and subsequently decreases with a further increase in $\beta$. This finding aligns with our analysis, as larger values of $\beta$ tend to remove more false positive tokens. However, excessively large values of $\beta$ can also result in the removal of true positive tokens, thereby reducing overall performance.


\begin{table}[t]
\centering
            \fontsize{9}{11}\selectfont
\setlength{\tabcolsep}{4mm}{
\begin{tabular}{lcc}
\toprule
Dataset & Constraint in CD & Ours \\
\midrule

CMeEE-V2     & 66.38  & \cellcolor{gray!10}\textbf{67.45} \\
CMeIE-V2     & 46.11  & \cellcolor{gray!10}\textbf{46.83} \\
IMCS-V2-NER  & 89.02  & \cellcolor{gray!10}\textbf{89.49} \\
CMedCausal   & 41.73  & \cellcolor{gray!10}\textbf{42.28} \\
IMCS-V2-SR   & 72.64  & \cellcolor{gray!10}\textbf{73.01} \\
CHIP-MDCFNPC & 42.88  & \cellcolor{gray!10}\textbf{43.71} \\
\bottomrule 
\end{tabular}}
\caption{Comparison of token constraint method on all datasets using ChatGLM-6B.}
\label{tab:comparisonConstraint}
\end{table}

\subsection{Comparison of Token Constraint} \label{sec:Constraint Methods}

To further validate the effectiveness of our proposed constraint method for avoiding noisy tokens in contrastive decoding, we compare against the constraint method of CD. Specifically, we replace the token constraint related to scale and range in ALCD with a constraint employed in CD, while maintaining the alternative contrastive decoding technique unchanged.
As shown in Table~\ref{tab:comparisonConstraint}, our method consistently outperforms the `constraint in CD' approach across all datasets. We attribute this improvement to the successful implementation of alternating adaptive token constraints on both scale and scope in our ALCD, whereas CD relies solely on a maximum value judgment.

\subsection{Affect of Decoupling Steps} 
\begin{figure}[t]
\centering
\includegraphics[width=0.9\columnwidth]{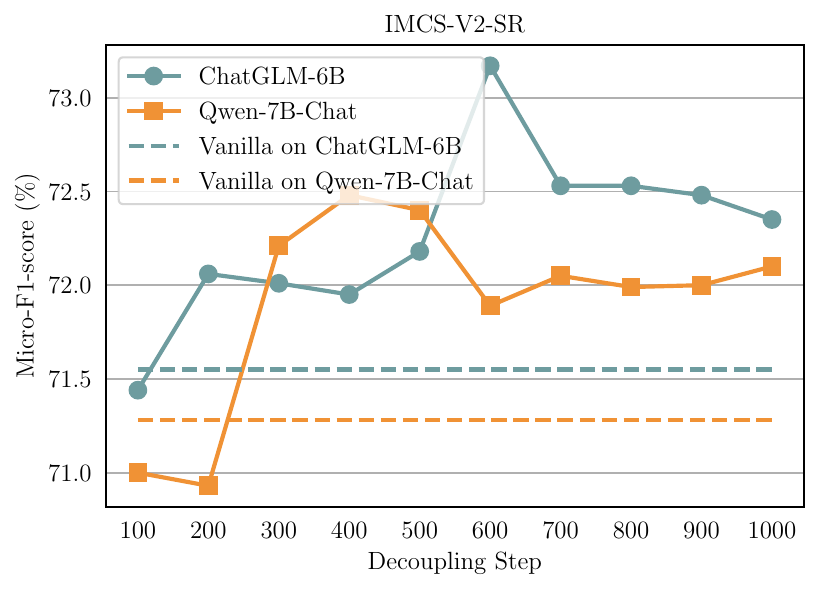}
\vspace{-0.5em}
\caption{Analysis of varying decoupling steps during fine-tuning on IMCS-V2-SR dataset. `Vanilla' refers to the performance of normal model using greedy search.
}
\label{fig:Steps}
\end{figure}

To investigate how the capabilities of sub-models affect overall performance of ALCD, we conducted experiments by individually fine-tuning two sub-task LLMs (i.e., $\mathcal{M}_{id}$ and $\mathcal{M}_{cl}$) with varying steps while keeping normal model (i.e., $\mathcal{M}_{nl}$) unchanged.
As illustrated in Figure~\ref{fig:Steps}, we observed that fine-tuning on sub-models effectively enhances performance, resulting in higher micro F1 scores compared to vanilla ones with 300 steps or larger.
When the number of fine-tuning steps increases, the performance rises for both LLMs, while decreases after 600 steps for ChatGLM-6B and 400 steps for Qwen-7B-Chat, respectively.
We believe the reason is that excessive fine-tuning steps can potentially improve the identification capabilities of $\mathcal{M}_{cl}$  and the classification capabilities of $\mathcal{M}_{id}$, consequently compromising the desired decoupling effect between the two abilities. As a result, contrasting the predictions in ALCD fails to improve performance.

\subsection{Case study}
We also present a case study from the CHIP-MDCFNPC dataset in Table \ref{tab:casestudy}. It can be observed that the results generated by gready search identify entities (e.g., `Swollen throat') that do not exist in the text or classify symptoms incorrectly (e.g., `classifying Cough as Other'). However, our method can accurately predict \textit{High fever: Positive, Cough: Positive, Fever: Positive, Common cold: Positive}. This further demonstrates the effectiveness of our approach in addressing hallucination issues.

\begin{CJK*}{UTF8}{gbsn}
\begin{table*}[h]
\centering
\setlength{\tabcolsep}{2mm}{
\begin{tabular}{p{3.5cm}p{11.5cm}}
\toprule
Input & \verb|请根据下面医生与患者之间的对话片段判断临床发现实体的阴阳性：\n患者：学校活动娃穿太少后感冒发烧，白天晚上反复发高烧都39度多！并且发烧多次～有偶尔有点咳嗽！发高烧都是喝的美林退烧。{口服的小儿芬黄颗粒，利巴伟。阿莫西林，止咳冲剂}快两天了这样高烧反复发作为什么还不好！求方法跪谢（男，8岁）\n医生：你好，宝宝除了发烧还有其他症状\n患者：偶尔带点咳嗽。|   \\
 (English Translation) & Please determine the positive and negative results of clinical findings based on the dialogue between the doctor and the patient below: Patient: After wearing too little clothing for school activities, the child caught a cold and had a high fever. The fever has been recurring during the day and night, reaching over 39 degrees Celsius each time. Occasionally, there is a slight cough. The fever is relieved by drinking Meilin. It has been almost two days, why is the high fever still not getting better? Please help (male, 8 years old). Doctor: Hello, besides the fever, does the baby have any other symptoms? Patient: Occasionally, there is a slight cough. \\
\hline
Gready search output & \verb|咳嗽：其他\n高烧：阳性\n喉咙肿大：阳性| \\
 (English Translation) & Cough: Other High fever: Positive Swollen throat: Positive\\
\hline
ALCD output (ours) &  \verb|高烧：阳性\n咳嗽：阳性\n发烧：阳性\n感冒：阳性| \\
(English Translation)& High fever: Positive Cough: Positive Fever: Positive Common cold: Positive \\
\bottomrule 
\end{tabular}}
\caption{Case Study in the CHIP-MDCFNPC dataset.}
\label{tab:casestudy}
\end{table*}
\end{CJK*}

%% file: section/conclusion.tex
\section{Conclusion} \label{sec:conclusion}
In this paper, we propose ALCD to address hallucinations of LLMs in MIE tasks.
ALCD utilizes decoupled fine-tuning process to separately learn LLM's identification and classification abilities. During inference, ALCD alternately enhances these abilities while excluding other capabilities that may result in hallucinations.
We also introduce adaptive scales based on distribution similarities to enable the flexible use of identification or classification abilities.
Extensive experiments conducted on two backbones have demonstrated substantial enhancement achieved by ALCD in MIE tasks.

\section{Limitation} \label{sec:limit}
Our approach aims to decouple the identification and classification abilities of LLMs in the medical information extraction tasks and leverage their respective capabilities through alternate contrastive decoding. However, this strategy leads to an increase in both fine-tuning and inference costs. In this paper, ALCD switches between identification or classification capabilities based on simple rule-based judgment, but it is worth exploring more automatic and flexible judgment methods in future work.
Furthermore, we have only investigated the effectiveness of our approach in medical information extraction tasks, and expanding our ALCD framework to other medical tasks, other domains, and other language settings is an avenue for future exploration. Exploring more robust decoupling methods and contrasting decoding techniques are also potential future research directions.

\section{Acknowledgement}
This work was supported in part by the grants from National Natural Science Foundation of China (No.U23A20319, 62222213, 62072423). This research was partially supported by Research Impact Fund (No.R1015-23), APRC - CityU New Research Initiatives (No.9610565, Start-up Grant for New Faculty of CityU), CityU - HKIDS Early Career Research Grant (No.9360163), Hong Kong ITC Innovation and Technology Fund Midstream Research Programme for Universities Project (No.ITS/034/22MS), Hong Kong Environmental and Conservation Fund (No. 88/2022), and SIRG - CityU Strategic Interdisciplinary Research Grant (No.7020046), and Tencent (CCF-Tencent Open Fund, Tencent Rhino-Bird Focused Research Program).

%% file: section/appendix.tex
\section{Appendix}
\label{sec:appendix}

\subsection{Tasks and Datasets} \label{appendix:task}
In the experiments, we adopt a Chinese medical dataset, named PromptCBLUE~\cite{zhu2023promptcblue}, including several common tasks. Due to limited resources, we select $6$ tasks for validation. The statistics are in Table~\ref{tab:Statistics}, and the dataset details are listed as follows:
\begin{itemize}
    \item \textbf{CMeEE-V2}. Chinese medical name entity recognition. We consider ``extracting entities from medical texts'' as \textit{identify} and ``categorizing the entities'' as \textit{classify}.
    \item \textbf{CMeIE-V2}. Chinese medical entity relation extraction. We consider ``recognizing the head and tail entities from medical texts'' as \textit{identify} and ``categorizing the relation types between entities''.
    \item \textbf{IMCS-V2-NER}. Medical entity recognition from the doctor-patient dialogue. We consider ``identifying the medical entities from dialogues'' as \textit{identify} and ``classifying the medical entity types'' as \textit{classify}.
    \item \textbf{CMedCausal}. Causal relation extraction for medical texts. We consider ``recognizing the causal and effect words from medical texts'' as \textit{identify} and ``categorizing the causal relation'' as \textit{classify}.
    \item \textbf{IMCS-V2-SR}. Medical normalization of the doctor-patient dialogue. We consider ``extracting the normalized words from dialogues'' as \textit{identify} and ``imputing the normalization labels'' as \textit{classify}.
    \item \textbf{CHIP-MDCFNPC}. Clinical concept finding and discrimination for the clinical report. We consider ``extracting the clinical concepts from reports'' as \textit{identify} and ``classifying the derived clinical concepts'' as \textit{classify}.
\end{itemize}

We show the number of prompt templates for each dataset \ref{tab:prompttemplate}. The large number of prompt templates in this PromptCBLUE allows us to better validate our ideas and the generalization ability of our model.
\begin{table}[t]
\centering
            \fontsize{9}{11}\selectfont
\setlength{\tabcolsep}{2mm}{
\begin{tabular}{lc}
\toprule
Variations & prompt templates   \\
\midrule
 CMeEE-V2 & 23 \\
 CMeIE-V2 & 37 \\
 CHIP-MDCFNPC & 14 \\
 IMCS-V2-NER & 25 \\
 IMCS-V2-SR & 13 \\
 CMedCausal &12 \\
 Overall & 124 \\
\bottomrule 
\end{tabular}}
\caption{The number of Prompt Templates for each dataset.}
\label{tab:prompttemplate}
\end{table}

\begin{table}[h]
\centering
            \fontsize{9}{11}\selectfont
\setlength{\tabcolsep}{2mm}{
\begin{tabular}{lcc}
\toprule
Variations & Beam search &  ours  \\
\midrule
Identification\_not\_exit     & 0.0767 & 0.0593 \\
Identification\_pred\_wrong  & 0.5241 & 0.5068\\
Classification\_pred\_wrong   &  0.0908 & 0.0775\\
\bottomrule 
\end{tabular}}
\caption{LLM's prediction accuracy in identification and classification.}
\label{tab:hallucination}
\end{table}

Furthermore, we provide some examples in Table \ref{tab:examples1} and \ref{tab:examples2}.

\begin{CJK*}{UTF8}{gbsn}
\begin{table*}[h]
\centering
\setlength{\tabcolsep}{2mm}{
\begin{tabular}{p{2cm}p{13cm}}
\toprule
IMCS-V2-SR & prompt template   \\
\midrule
  template1 & \verb|找出当前对话中的症状，并判断阴阳性：\\n[INPUT_TEXT]\\n症状阴阳性选项：[LIST_LABELS]\\n答：| \\
 (English translation) & Identify the symptoms in the current conversation and determine their positive or negative nature: [INPUT\_TEXT] Symptom positive/negative options: [LIST\_LABELS] Answer:  \\
\hline
 template2 & \verb|[INPUT_TEXT]\\n根据上述对话历史，当前对话中症状有哪些？这些症状的阴阳性是？\\n选项：[LIST_LABELS]\\n答：|  \\
 (English translation) & [INPUT\_TEXT] Based on the previous conversation history, what are the symptoms in the current conversation? What is the positive or negative nature of these symptoms? Options: [LIST\_LABELS] Answer: \\
\hline
 template3 & \verb|依据之前对话内容，抽取当前对话中出现的症状实体，并明确它们的阴阳性：\\n[INPUT_TEXT]\\n备选阴阳性标志：[LIST_LABELS]\\n答：|\\
  (English translation) & Based on the previous conversation content, extract the symptom entities that appear in the current conversation and specify their positive or negative nature: [INPUT\_TEXT] Available positive/negative indicators: [LIST\_LABELS] Answer: \\
\bottomrule 
\end{tabular}}
\caption{Examples in the IMCS-V2-SR dataset.}
\label{tab:examples1}
\end{table*}

\begin{table*}[]
\centering
\setlength{\tabcolsep}{2mm}{
\begin{tabular}{p{2cm}p{13cm}}
\toprule
CMeIE-V2 & prompt template   \\
\midrule
  template1 & \verb|找出指定的三元组：\\n[INPUT_TEXT]\\n实体间关系：[LIST_LABELS]\\n答：| \\
(English translation) & Find the specified triplet: [INPUT\_TEXT] Relationship between entities: [LIST\_LABELS] Answer:   \\
\hline
 template2 & \verb|根据下述文本，提取出具有[LIST_LABELS]关系的实体对：\\n[INPUT_TEXT]\\n答：|  \\
 (English translation) & Based on the following text, extract entity pairs with the relationship [LIST\_LABELS]: [INPUT\_TEXT] Answer:  \\
\hline
 template3 & \verb|[INPUT_TEXT]\\n问题：找出句子中描述的[LIST_LABELS]三元组的内容。\\n答：|\\
  (English translation) & [INPUT\_TEXT] Question: Find the content of the triplet described in the sentence for [LIST\_LABELS]. Answer: \\
\bottomrule 
\end{tabular}}
\caption{Examples in the CMeIE-V2 dataset.}
\label{tab:examples2}
\end{table*}
\end{CJK*}

\subsection{Ability to Mitigate Hallucinations}
Furthermore, we conducted experiments to validate our method's ability to mitigate hallucinations. We evaluated the LLM's prediction accuracy in identification and classification. This includes three categories:
  \textbf{Identification\_not\_exist}: The entity or symptom identified by the LLM does not exist in the original text.
      \textbf{Identification\_pred\_wrong}: The entity or symptom identified by the LLM exists in the original text, but it is incorrect.
      \textbf{Classification\_pred\_wrong}: The LLM correctly identifies the entity or symptom, but the classification result is incorrect.
Table \ref{tab:hallucination} shows the results based on Qwen and CHIP-MDCFNPC datasets.

The table presents the results indicating that the effectiveness of ALCD is enhanced when evaluating classification and recognition capabilities separately. In all three categories, our ALCD method achieved lower error rates compared to the Beam Search method. This suggests that our method has some effectiveness in reducing LLM hallucinations. For the "Identification\_not\_exist" category, the ALCD method is more accurate in identifying entities or symptoms and generates less content that does not exist in the original text. For the "Identification\_pred\_wrong" category, which refers to cases where the model identifies entities or symptoms that exist in the original text but makes errors in identification, it indicates that the ALCD method has improved in correctly identifying entities or symptoms. For the "Classification\_pred\_wrong" category, it suggests that the ALCD method has also improved in classifying entities or symptoms.
This demonstrates that ALCD has been proven to effectively minimize specific types of errors (identification and classification).